\documentclass[letterpaper, 10 pt, conference]{ieeeconf}  
\IEEEoverridecommandlockouts                              
\usepackage{graphics}    
\usepackage{times}       
\usepackage{amsmath}     
\usepackage{amssymb}     
\usepackage{graphicx}
\usepackage[noend]{algpseudocode}
\usepackage{blindtext}
\usepackage{subfig}
\usepackage{hhline}
\usepackage{graphicx}
\usepackage{subfig}
\graphicspath{{images/}}
\usepackage{times}    
\usepackage{svg}
\usepackage{amsmath}     
\usepackage{amssymb}     
\usepackage{graphicx}
\usepackage[noend]{algpseudocode}
\usepackage{url}
\usepackage{amsmath}
\usepackage[linesnumbered,ruled]{algorithm2e}
\usepackage{siunitx}
\usepackage{colortbl}
\let\labelindent\relax
\usepackage{enumitem}
\usepackage{cite}
\usepackage{float}

\usepackage[font=small]{caption} 
\usepackage{tabularx}
\usepackage{makecell}
\usepackage{multirow}
\usepackage{hhline}
\usepackage{tikz}
\usetikzlibrary{backgrounds, fit, shapes.geometric, positioning}
\usepackage[noend]{algpseudocode}
\usepackage{mathtools}
\usepackage{amssymb}
\usepackage{float}

\newcolumntype{Y}{>{\centering\arraybackslash}X}
\usepackage[font=small]{caption}

\def\secref#1{Sec.~\ref{#1}}

\def\tabref#1{Tab.~\ref{#1}}
\def\eqref#1{Eq.~(\ref{#1})}

\newcommand\etal{\emph{et al.}}

\newcommand*{\algotitle}[2]{%
  \stepcounter{algocf}%
  \hypertarget{algocf.title.\theHalgocf}{}%
  \NR@gettitle{#1}%
  \label{#2}%
  \addtocounter{algocf}{-1}%
}

\usepackage{makecell}
\usepackage{footmisc}
\usepackage{tabu}

\makeatletter
\newcommand{\mathleft}{\@fleqntrue\@mathmargin0pt}
\newcommand{\mathcenter}{\@fleqnfalse}
\makeatother

\title{Viewpoint Push Planning for Mapping of Unknown Confined Spaces}

\author{Nils Dengler \and Sicong Pan \and Vamsi Kalagaturu \and Rohit Menon \and Murad Dawood \and Maren Bennewitz
  \thanks{All authors are with the Humanoid Robots Lab, University of
    Bonn,  Germany. Murad Dawood and Maren Bennewitz are additionally with the Lamarr Institute for Machine Learning and Artificial Intelligence, Germany.
 		This work has partially been funded by the European Commission
  		under grant agreement number 964854 --RePAIR --
                H2020-FETOPEN-2018-2020 and
                by the Deutsche Forschungsgemeinschaft (DFG, German
                Research Foundation) under Germany's Excellence
                Strategy, EXC-2070 -- 390732324 -- Phenorob and under the grant number BE 4420/4-1 (FOR 5351: KI-FOR Automation and Artificial Intelligence for Monitoring and Decision Making in Horticultural Crops, AID4Crops).} 
}

\begin{document}
\maketitle
\thispagestyle{empty} 
\pagestyle{empty}

\begin{abstract} Viewpoint planning is an important task in any application where objects or scenes need to be viewed from different angles to achieve sufficient coverage.
The mapping of confined spaces such as shelves is an especially
challenging task since objects occlude each other and the scene can
only be observed from the front, posing limitations on the possible viewpoints.
In this paper, we propose a deep reinforcement learning framework that generates promising views aiming at reducing the map entropy.
Additionally, the pipeline extends standard viewpoint planning by predicting adequate minimally invasive push actions to uncover occluded objects and increase the visible space.
Using a 2.5D occupancy height map as state representation that can be efficiently updated, our system decides whether to plan a new viewpoint or perform a push.
To learn feasible pushes, we use a neural network to sample push
candidates on the map based on training data provided by human experts. 
As simulated and real-world experimental results with a robotic arm show, our system is able to significantly increase the mapped space compared to different baselines, while the executed push actions highly benefit the viewpoint planner with only minor changes to the object configuration.
\end{abstract}

\section{Introduction}
\label{sec:intro}

Viewpoint planning (VPP) is an important part of various robotic applications to gain an understanding of the environment.
For example, it enables a robot to identify occluded objects on a crowded table \cite{li2021robotic} or to estimate the amount of harvestable fruits in horticulture~\cite{zaenker2021viewpoint}.
For manipulation purposes, VPP can also output the best grasping position by viewing and mapping the object from different angles~\cite{breyer2022closed}.
However, the performance of VPP is highly dependent on the given scenario.
While the possibility of good viewpoints, and thus scene coverage, increases when the robot is able to act freely in the environment, the environment itself can constrain actions and reduce the possible coverage.
A confined indoor space such as a shelf or refrigerator is an example of such a constrained environment.
If the robot's task is to map the current state of a shelf, the difficulty increases with the number of objects present and their closeness.
Such constrained scenes can mainly be viewed from one direction, i.e.,
the front with a limited number of possible viewpoints.

\begin{figure}[t]
  \centering
 \includegraphics[width=0.99\linewidth]{"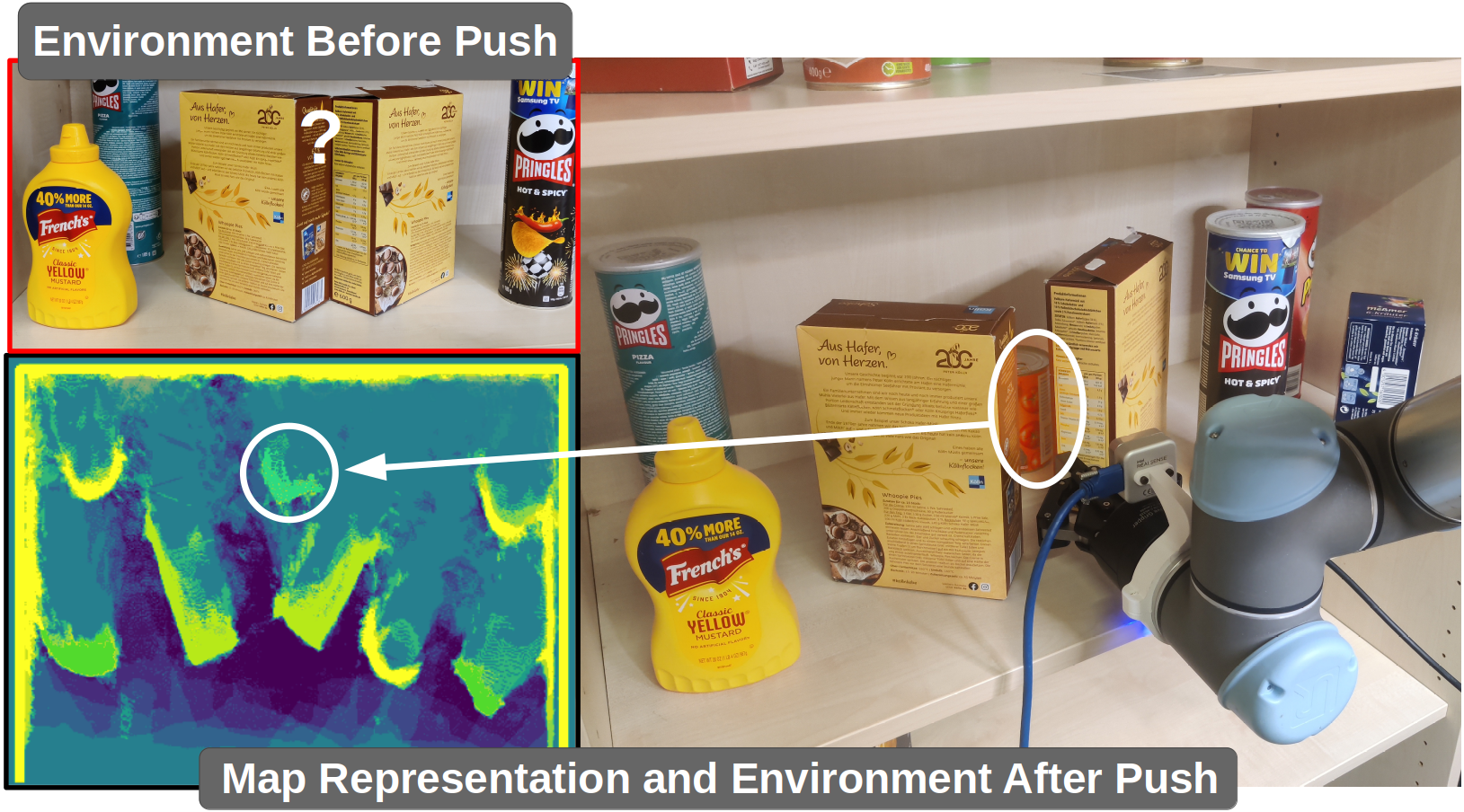"}
 \caption{Motivation of our approach. The robot's task 
   is to map the shelf board as completely as possible in a 2.5D
   occupancy height map. Without manipulative actions it is not
   possible to plan viewpoints necessary to uncover spaces behind big
   objects. However, with just a short push of the right box, the
   information value of the same viewpoint increases because of the
   newly uncovered object~(white circle).}
  \label{fig:motivation}
\end{figure}

In particular, the limited top-down view from above the scene makes it difficult to identify and map objects that are hidden behind other objects or closer to the back of the shelf.
An obvious solution to this problem is to introduce manipulation actions to uncover previously occluded space.
However, human preferences play an important role in the arrangement of indoor environments~\cite{abdo2015robot}.
Therefore, not all manipulation actions are equally suitable, as they could lead to unwanted large changes in the layout of the scene.
To overcome this problem, minimally invasive pushing actions should instead be used to uncover previously occluded areas.
Furthermore, in cluttered scenes it is easier to identify promising non-prehensile push positions than, for example, grasping poses on objects to rearrange them, since the constrained environment limits the number of possible actions.

Fig.\ref{fig:motivation} shows an example scenario where a packed shelf board needs to be mapped.
As can be seen, a short minimally invasive push is sufficient to move one of the larger boxes, exposing the space behind it and uncovering the previously occluded can.

In this paper, we therefore present a novel framework called viewpoint push planning.
We apply deep reinforcement learning (DRL) to perform VPP in confined scenarios, while predicting suitable minimally invasive, non-prehensile push actions to improve shelf coverage.
In particular, for the immense number of possible object configurations within the shelf, DRL can learn the best viewpoints from experience and generalize them to different configurations.
Furthermore, we use a push prediction network trained in a supervised manner that outputs the best pushing action to free views and aid the mapping process.
As representation of the environment, our framework generates a
2.5D~occupancy height map fron 3D point clouds obtained with an RGB-D
images of a camera mounted on the robot's end effector.
As long as the VPP finds a new promising viewpoint that increases the map coverage, it is not necessary to perform a push action, as it would unnecessarily alter the environment. 
Therefore, we use an action selection method to decide whether a VPP or a pushing action is best given the current situation.

As shown in our experimental evaluation, our learned VPP is able to significantly reduce the entropy of the map compared to two baselines and two VPP systems from the literature~\cite{pan2023global,delmerico2018comparison}.
The complete framework, including the execution of push predictions, is able to reduce the entropy even further.
The main contributions of our work are as follows: \begin{itemize} 
\item A 2.5D heightmap representation for mapping of confined spaces.
\item A viewpoint planning framework including pushing actions 
based on deep reinforcement learning.
\item A qualitative and quantitative evaluation in simulation as well as on a real robot including the usability of our map representation for object reasoning and retrieval tasks.
\end{itemize} 

\begin{figure*}[th!]\centering \includegraphics[width=0.8\textwidth]{"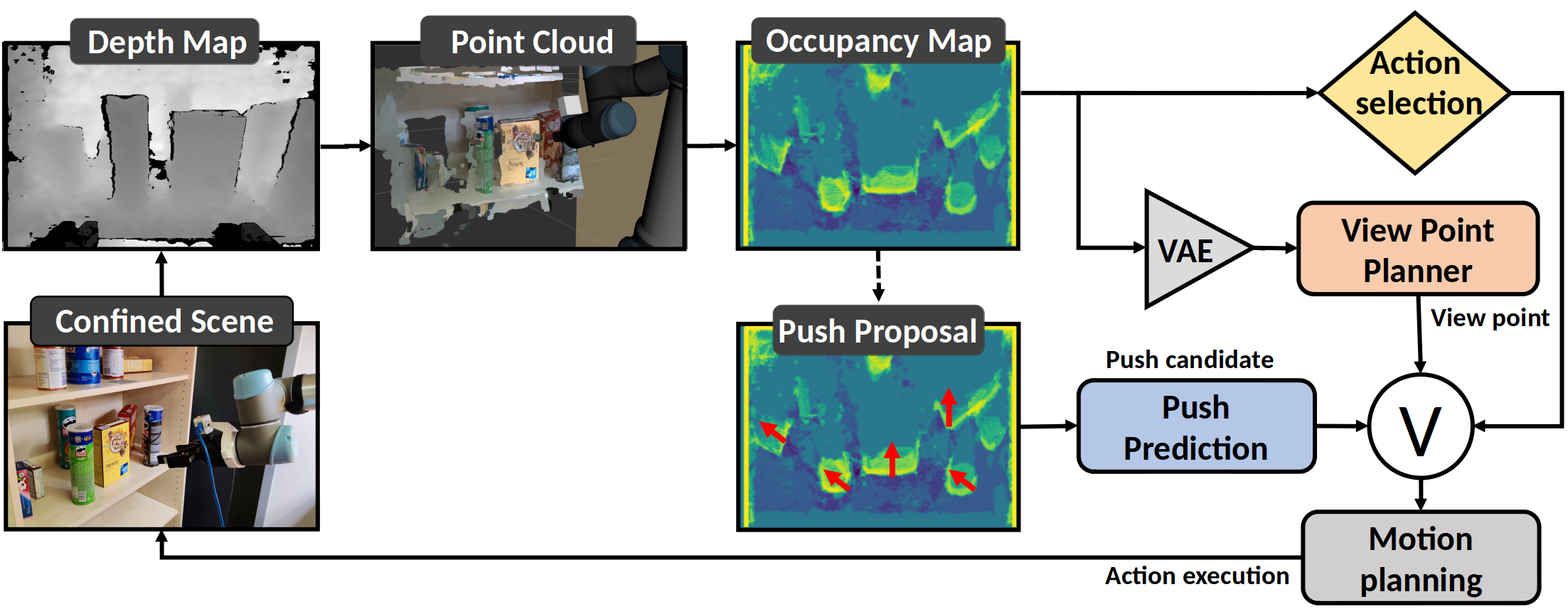"} \caption{Overview of our viewpoint pushing framework.
Our system receives a depth map from an RGB-D camera on the end-effector of the robotic arm and converts it into the corresponding point cloud.
We then orthographically backproject the point cloud into a 2.5D occupancy height map and feed it into a variational auto-encoder~(VAE) to obtain the latent space.
An action selection method decides whether to map or to perform a predicted push or the next viewpoint action.
The latent space is used for the viewpoint planner to select the next best viewpoint, while push proposals sampled from the height map are used to predict the best push candidate.}
\label{overview}
\end{figure*} 

\section{Related Work}
\label{sec:related}

Viewpoint planning has been used in applications ranging from the reconstruction of single objects to large-scale applications such as the mapping of crops in green houses.
While in the past, sample-based approaches were mostly used to perform viewpoint planning~\cite{zaenker2021viewpoint,lehnert20193d, scott2003view, morrison2019multi}, recently learned planners have become more popular~\mbox{\cite{hepp2018learn, wang2019autonomous, pan2022scvp}}.
However, reinforcement learning, as in this paper, has not often been
used for VPP, although it can highly benefit VPP approaches by reducing the planning time and learn a
generalizable planning behavior for multiple objects and scenes.
Peralta~\etal~\cite{peralta2020next} applied reinforcement learning to generate best views for the reconstruction of 3D~houses and reduced the number of needed planning steps, while increasing the mapping accuracy wrt.
a uniformly sampled approach.
Zeng~\etal~\cite{zeng2022deep} used reinforcement learning to map sweet pepper crops with a robotic arm and an in-hand camera in a greenhouse environment.
The authors applied an octomap and ray casting to determine the next observation for the agent.
However, this increases the time needed to plan an action compared to our approach, which relies on a 2.5D representation.


Previous work in the area of interaction with objects in confined spaces has concentrated on reasoning about individual objects~\cite{huang2021mechanical, huang2022mechanical, huang2023mechanical} or their retrieval~\mbox{\cite{bejjani2021occlusion, kang2022grasp, ahn2022coordination, nam2020fast, li2016act}}.
In contrast to our work, most of these approaches use fixed cameras, no long-term updated map representation, or assume that a lot of object information is known a priori.
Furthermore, none of these approaches consider the invasiveness of the actions and perform each action regardless of the resulting scene change.
For example, Bejjani~\etal~\cite{bejjani2021occlusion} used a single-shot representation and reinforcement learning for object retrieval with initially unknown object positions.
Due to the incomplete map representation, the agent might investigate irrelevant areas first, while possibly changing the layout of objects completely.
Using a complete map for action planning would counteract this behavior.
\mbox{Huang~\etal~\cite{huang2021mechanical}} rely on a fixed camera position for object reasoning with Gaussian distributions for likely positions of an occluded object.
By using different camera positions and VPP, as in this paper, some objects would become visible even though they are occluded from the front, which supports object reasoning.

There has not been much research regarding the long-term mapping of confined spaces.
Miao~\etal~\cite{miao2022safe} recently proposed a system to reconstruct objects placed in shelves.
The authors create an occlusion dependency graph to reason about the relations between the objects and to calculate the best retrieval and rearrangement strategy.
However, the authors assume that initial object models are given, which is a strong assumption.
Furthermore, the complete 3D reconstruction of objects is time-consuming and not the focus of this work.
While the reconstructed object knowledge can also be used to reduce planning time for object reasoning and retrieval, our approach is faster, since not all objects need to be processed individually.

\section{Our Approach}
\label{sec:main}
In the following, we first give an overview of our system and introduce our framework in detail afterwards.

\subsection{Overview}
We consider the following problem.
In the confined environment of a shelf, a stationary robotic arm needs to map a shelf board as completely as possible.
The shelf board is packed with a number of different objects, which may occlude each other.
A schematic overview of our approach is shown in Fig.
\ref{overview}.
Our system receives a depth map of the environment from an RGB-D camera mounted on the end-effector of the arm and orthographically back-projects it from the corresponding point cloud into a 2.5D occupancy height map.
In order to decouple the feature extraction from the learning process and to promote faster convergence of the reinforcement learning agent for viewpoint planning, we use a variational auto-encoder \cite{kingma2013auto} to encode the height map into a latent space of size 32.
Since object occlusion can lead to areas that cannot be fully mapped by viewpoint planning alone, we sample push proposals on the height map to predict the best push candidate that will uncover additional space in a minimally invasive manner.
Depending on the change in mapped space, an action selection method decides whether the best action in this situation is a push, moving to a new viewpoint, or if the mapping process has finished.

\subsection{Environment Representation}

\label{section:representation}
\begin{figure}[t]
  \centering
  \subfloat[\centering Occupancy map]{{\centering\includegraphics[width=0.5\linewidth]{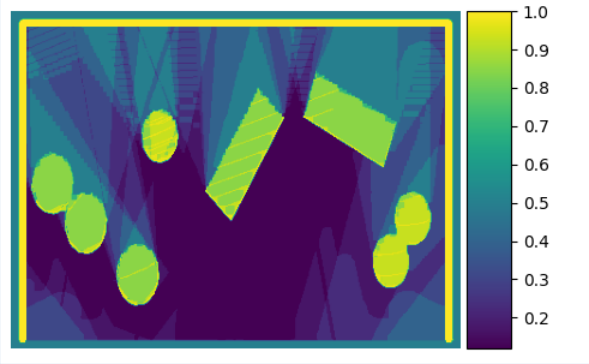} }}%
    \subfloat[\centering Height map]{{\centering\includegraphics[width=0.5\linewidth]{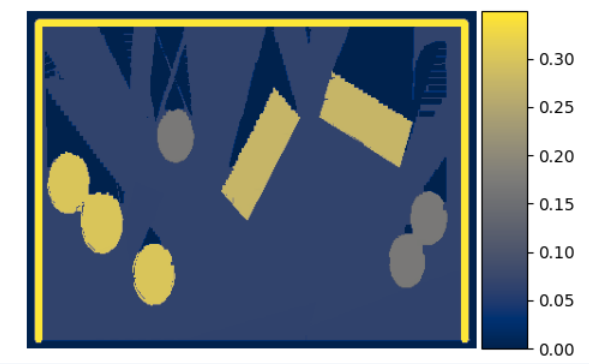} }}%
 \caption{Example visualization of the 2.5D state representation after mapping. The left image shows occupancy probability values of the map, while the right image shows the maximum height value of each cell. The brighter the color of the map, the higher its value.}
  \label{fig:heightmap}
\end{figure}

We use a 2.5D representation of the environment since it can be efficiently updated.
From a depth image, we compute the corresponding point cloud and orthographically back-project it into the top-down view of the scene 
to calculate a 2.5D occupancy height map as shown in Fig.\ref{fig:heightmap}.
To update the map, we implemented \mbox{log-odds} for occupancy updates~\cite{moravec1985high} and update only the cells in the current field of view.
For the height component of the map, we use the maximum measured height value for each cell.

Note that In case grasping requires a more accurate representation of the objects, our map can be used to plan a path to the object, while the actual grasping action can be performed on a local higher dimensional representation.
We provide an application example of our map for object reasoning and grasping in the experimental section.
As previous approaches have shown, 2.5D representations can be used to compute grasping points and manipulate objects~\cite{zeng2018learning, mohammed2020pick, zeng2020tossingbot}.


\subsection{Viewpoint Planning (VPP)}
\label{sec:VPP}
For VPP, we use reinforcement learning (RL) to generate the next best view.
The overall goal of our VPP agent is to map the environment as completely as possible.
As in previous approaches \cite{hepp2018plan3d}, we generate three fixed viewpoints at the beginning of each episode, which the agent uses to generate an initial map.
This reduces the need for the agent to learn good initial viewpoints that are similar for each state configuration.
For the three fixed viewpoints, we use a camera pose facing towards the center of the shelf and two poses at the left and right corners of the shelf, pointing diagonally inside the shelf to the opposite corner.
By using these three viewpoints, we usually achieve an initial map coverage of around 50\%.

To evaluate the quality of the agent during testing and for training, we use the entropy $h(M)$, information gain $IG(M)$, and motion cost $c(p_t)$ as defined below: 
\begin{equation}
h(M)= \frac{m_u^t}{cells_M}
\end{equation}
\begin{equation}
IG(M) = \frac{m_u^{t-1} - m_u^{t}}{m_u^{t-1}}
\end{equation}
\begin{equation}
c(p_t) = d(p_{t-1}, p_t )
\end{equation}
\label{vpp_equ}
$h(M)$ states the ratio of the unknown cells $m_u^t$ to the total number of cells of map $M$ at time~$t$.
The lower the entropy, the higher the information value of the map.
Note that we consider each cell with an occupancy probability value $\mathit{occ} > 0.5+\tau_{\mathit{unknown}}$ as occupied, $\mathit{occ} < 0.5 - \tau_{\mathit{unknown}}$ as free, and unknown otherwise.
$IG(M)$ indicates the percentage decrease of unknown cells $m_u$ from the previous time step $t-1$ to time step $t$.
Finally, we use the Euclidean distance $c(p_t)$, between two viewpoints $p_{t-1}$ and $p_t$ to approximate the move cost.

During training, an episode ends if the change of $h(M)$ for the last three viewpoints is below a certain threshold~$\tau_{\mathit{single}}$ and the sum of changes is below the threshold $\tau_{\mathit{sum}}$.
Furthermore, we terminate the episode if any part of the arm has contact with an object or an object drops in or out of the shelf.

In the following, we define the actions, observations, and the reward function of our agent.

\subsubsection{\textbf{Action Space}} As action we use the 5D Cartesian pose~$(x, y, z, pitch, yaw)$ of the camera on the end effector.
We ignore the roll, as it does not have much impact on the information gain of a viewpoint.
To speed up training and reduce the likelihood of damage to the arm, we restrict the $(x,y,z)$~space of the actions to a reasonable area in front of the shelf with a small overlap inside the shelf.
For pitch and yaw, we experimentally evaluated the angle of rotation so that the arm cannot damage the camera during its movements.


\subsubsection{\textbf{Observation Space}} 
The main part of the observation space is the latent space of the state representation of size~32.
In addition, we provide the last action in 5D Cartesian coordinates as well as the information gain the agent receives from this viewpoint and the motion cost to move there.
Furthermore, we also included a Boolean collision indication.
Finally, we calculate from our map the center point of the largest unknown area and provide it to the agent as 3D Cartesian coordinates to give the agent an indication of a promising region of interest.
 
\subsubsection{\textbf{Reward}} 
\label{vpp:reward}
We keep the reward function as simple as possible to aid the training.
The reward consists of two parts: \mathleft 
\begin{equation}
r_{\mathit{coll}}= 
\begin{cases}
-\eta,\hspace{.2cm} \text{if collision or drop from shelf}\\
0,\hspace{.2cm} \text{else}
\end{cases}
\end{equation}
\mathleft 
\begin{equation}
r_{\mathit{gain}}= \hspace{.5cm} \alpha * - c(p_t) + \beta * IG(M)
\end{equation}
$r_{\mathit{coll}}$ penalizes each contact with the object according to a predefined high negative value $\eta$.
Since the goal is to map the environment in as few time steps as possible, we penalize each step in $r_{\mathit{gain}}$ according to the move cost $c(p_t)$.
The higher $IG(M)$, the more space has been uncovered during one step, which is positively rewarded.
$\alpha$ and $\beta$ are used to weight closer viewpoints with the effectiveness of the viewpoint.
Finally, the total reward can be expressed as $r = r_{\mathit{coll}} + r_{\mathit{gain}}$.

%
\subsection{Push Prediction}
\label{sec:pushpred}

\begin{figure}[t]
  \centering
{{\includegraphics[width=0.45\linewidth]{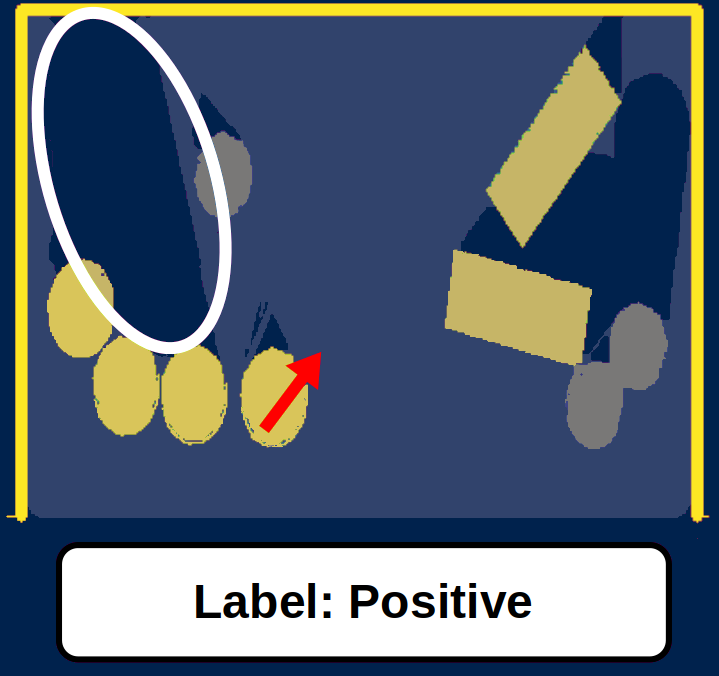} }}
    \centering {{\includegraphics[width=0.45\linewidth]{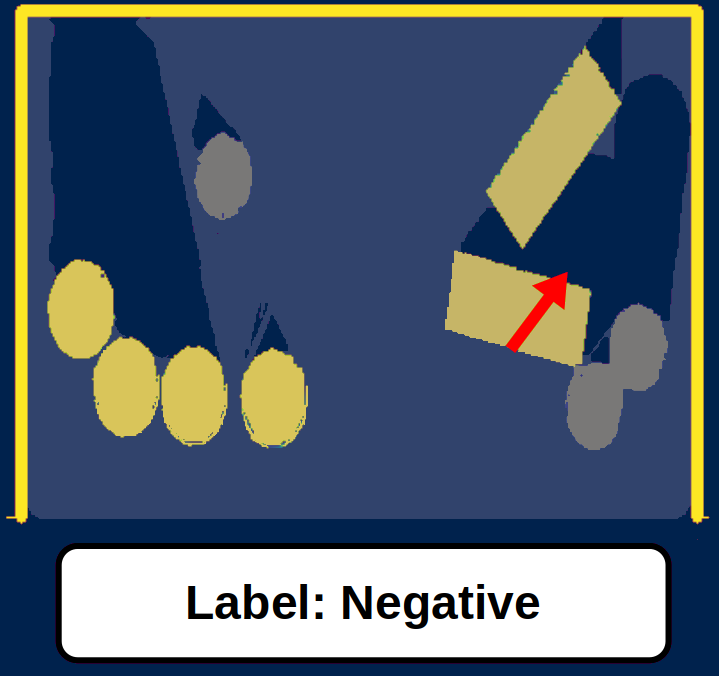}} }%
 \caption{Two height map object configurations with example sampled
   push candidates (red arrow). The brighter the color, the higher the
   object. Unknown space is marked dark blue. The labels indicate how the
   human experts evaluated the quality of the push candidate in terms
   of minimally invasiveness and additional uncovered space.}
  \label{fig:push_samples}
\end{figure}

Standard viewpoint planning performs actions aiming at minimizing the entropy without altering the environment.
In contrast, interactive perception \cite{bohg2017interactive} as proposed in this paper performs additional actions to change the current object configuration to support perception and coverage of the environment.
However, any manipulation action constitutes a trade-off between reducing the entropy of the environment and disrupting the current configuration of the scene.
Thus, we focus on minimally invasive actions to achieve a balance between these competing requirements.
In particular, we use non-prehensile push actions since they are easy to compute and carry out in confined spaces, while other types of manipulation actions would increase the computational efforts due to the constraints of the environment.

\subsubsection{Learning of the Push Prediction Network} To predict the push that best supports the VPP, we apply a similar approach as Eitel~\etal~\cite{eitel2020learning} to learn a prediction network that outputs the best push of the current scene.
Unlike the original approach, our goal is not to separate objects and thus change the configuration as much as possible, but to predict the minimally invasive push action that increases the area that can be mapped by the VPP agent, while balancing the trade off mentioned above.
We had three human experts manually label 5,000 simulated map configurations and an additional 850 configurations of the real-world scene.
The experts were instructed to label the images annotated with the starting point of the push and its direction, with binary labels "positive"$=1$ and "negative"$=0$.
Furthermore, they were told to consider the trade off between entropy
reduction and change in the configuration of the scene.
and prefer pushes that only minimally change the object configuration
while achieving a large entropy reduction. The maximum considered push length was 10 cm.

Example images including the labels are shown in Fig.~\ref{fig:push_samples}.
In the left image, the push candidate was labeled "positive" since the push would only change the position of one object but uncover the previously hidden space marked with the white circle.
In contrast, the second example was labeled "negative" because the push involves the displacement of up to four objects, which would be pushed against the edges of the shelf, potentially damaging them.

Note that we used manual rather than automatically generated labels because without computationally expensive physics simulations, it is difficult to automatically calculate the effect that pushing of an object may have on the object itself as well as on its neighbours.
Instead, we rely on human expertise.

\subsubsection{Generation of Push Proposals} 
\label{sec:data}
To generate push proposals, we sample a number of push candidates for each map configuration by ray casting from three fixed positions outside the shelf to the back of the shelf.
The 3D Cartesian coordinate of the first occupied cell along the ray is then considered a push candidate, along with eight push angles and a predefined push length.
In our experiments we sampled 100 to 150 push proposals on average.
As input to the network, we generate push maps by translating the starting point of the push to the center of the map and rotating the image according to the push direction.

Our prediction network consists of five convolution layers, each followed by a max-pooling layer, and the rectified linear unit activation function.
With the labeled push maps, we followed the training structure of \cite{eitel2020learning}, with the Adam optimizer and a learning rate of $1e-4$.
For inference, we feed the push proposal maps into the network and execute the push with the highest output of the network.

\subsection{Action Selection}
\label{selection}
Since a push is not required after each executed viewpoint, and pushing can lead to several new viewpoints that increase the observable space, viewpoint planning is always performed until the VPP RL agent terminates.
To decide afterwards whether another push is needed or the mapping process is finished, we proceed as follows.
We assume that the VPP agent is trained to an optimal policy and only terminates according to the criteria defined in \ref{sec:VPP}.
Furthermore, we assume that the push prediction network always outputs
the best possible push action. Since we assume the push prediction to
be optimal, if the previous push action did not lead to an entropy
reduction of $\Delta h(M) > \tau_{\mathit{push}}$, we do not perform
any further push action and terminate the mapping process. In this
way, we deal with the trade-off between
entropy reduction and change in the current configuration of
the scene and also avoid infinite pushing loops.

\section{Experimental Evaluation}
\label{sec:exp}
The goal of the experiments is to show the improved mapping result of our framework.
Therefore, we evaluate the entropy reduction, the number of steps needed, the planning time, and the object displacement of our complete pipeline and compare the performance to different baselines in simulation as well as in real-world experiments.
Furthermore, to show that our 2.5D map is a sufficient state representation of the environment, we demonstrate its application to object reasoning and grasping.

For the simulation, we use Pybullet \cite{coumans2021} and for the real-world experiments ROS~\cite{ros}.
Both experiments are performed on a 6 degree of freedom UR5\footnote{https://www.universal-robots.com/products/ur5-robot/} equipped with a Robotiq 2f85 gripper\footnote{https://robotiq.com/products/2f85-140-adaptive-robot-gripper}.
As camera, we use a Realsense D405 for close-range depth images mounted on top of the end effector.
We trained and evaluated our approach on a computer with an i7-6800k CPU and a RTX2070 GPU.
For the reinforcement learning agent we used the stable-baselines3 framework \cite{stable-baselines3} with the truncated quantile critics algorithm (TQC) \cite{kuznetsov2020controlling}.
Furthermore, we used the same actor-critic architecture with a small three-layered network of size \{256, 256, 256\} each.
To execute the actions we use pure OMPL \cite{sucan2012ompl} in simulation and OMPL with MoveIt \cite{coleman2014reducing} for the real robot.
For our experiments, we use a shelf with size of 40\,cm $\times$ 80\,cm
$\times$ 40\,cm. 
The reinforcement learning agent was trained on 150,000 iterations including approximately 25,000 shelf configurations. For each shelf configuration, we randomly sampled the number and position of the objects on the shelf. 

For the experiments, we used the following thresholds: $\tau_{\mathit{unknown}}=0.2$, $\tau_{\mathit{single}}~=~0.01$, $\tau_{\mathit{sum}}~=~0.05$, and $\tau_{\mathit{push}}~=~0.01$.
Furthermore, for the reward function of the VPP agent we used $\alpha = 1$ and $\beta=10$.
The implementation of our learning pipeline, both in simulation and on the real robot is available on Github\footnote{https://github.com/NilsDengler/view-point-pushing \label{github_code}}.

\subsection{Evaluation of the Viewpoint Planner}
\label{sec:eval_vpp}

\begin{table}
\normalsize
\centering
{
\begin{tabular}{|c|c|}
\hline
\textbf{Simulation} & \textbf{Entropy Reduction}\\ \hhline{|=|=|}
Ours  &  $\mathbf{30.0\%} \pm  0.9\%$\\ \hhline{|-|-|}
Random &  $15.5\% \pm  0.6\%$\\ \hhline{|-|-|}
RSE\cite{delmerico2018comparison} & $16.5\% \pm  0.5\%$\\ \hhline{|-|-|}
GMC \cite{pan2023global} & $21.8\% \pm  0.8\%$\\ \hhline{|=|=|}
\textbf{Real World} &\textbf{Entropy Reduction}\\ \hhline{|=|=|}
Ours &  $\mathbf{26.1\%} \pm  1.0\%$\\ \hhline{|-|-|}
Random &  $20.3\% \pm  1.0\%$\\ \hhline{|-|-|}
\end{tabular}
}
\caption{Evaluation of the viewpoint planner: Reduced entropy
  compared to the initial entropy of the map computed with 3P~(three
  fixed viewpoints). 
  Each metric is shown with its standard error.
  The experiments were carried out in 100 simulated and 15 real-world scenarios.
  As can be seen, our approach lead to the highest reduction in
  entropy in comparison to all baselines. The reduction is significant according to the paired t-test with a p-value of 0.05.}
\label{tab:VPP-Result}
\end{table}

To evaluate our viewpoint planner individually, we executed the 3-point fixed planner (3P) used for training and evaluated how much our learned planner improves the resulting map.
Furthermore, we compared it to a random method and reimplemented two sample-based state-of-the-art next-best view planners, a greedy planner by Delmerico~\etal\cite{delmerico2018comparison} (RSE) and a global planner by Pan~\etal\cite{pan2023global} (GMC).
We compare against RSE to show the limitation of greedy planners in confined spaces and against GMC to demonstrate the overall benefit of a learned approach over a sample-based method.
We adapted the methods to work also in our confined-space multi-objects scenario by sampling only reachable poses a priori and removing the region of interest constraint.
For comparability, the two state-of-the-art systems and the random sampling method used the same workspace as our system to generate the viewpoints, the map generated by 3P as initial representation, and the same number of view poses as our agent in the environment.

For the experiments, we sampled 100 different shelf configurations in simulation and 15 in real world, each with 8 to 10 objects.
As can be seen in Tab.~\ref{tab:VPP-Result}, our agent outperformed all baselines in simulation, significantly according to the paired t-test,  with an on average 30\% reduced entropy compared to 3P, while needing 4.94 viewpoints.
Since RSE is a greedy sampling-based method, it sometimes performed worse or equal to the random agent and therefore shows no large difference since it is designed for single object reconstruction and sticks to the local optimum.
The performance of GMC, in contrast, is better due to its global 
sampling, still our approach achieves a 37\% higher entropy reduction relative to GMC. 
Furthermore, GMC took 7.29\,s per planning step, whereas our approach planned the next viewpoint in 0.04\,s.

As can be seen in the lower part of Tab.~\ref{tab:VPP-Result}, the simulation results are confirmed also during the real-world experiments, where our system was able to perform almost as well as in simulation.

\subsection{Evaluation of the Viewpoint Push Planner}

\begin{figure}[t]
  \centering
  \subfloat[\centering Before pushing]{{\includegraphics[width=0.45\linewidth]{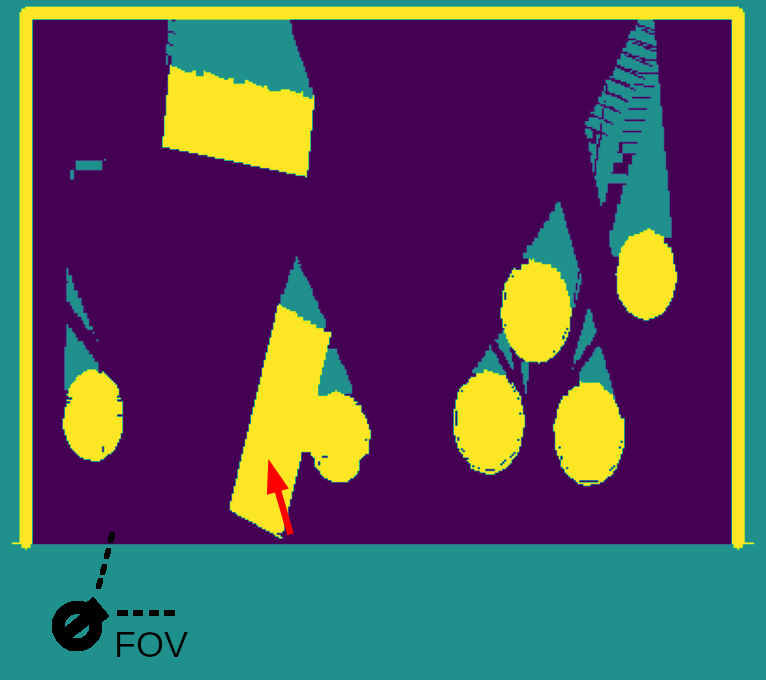} }}%
    \subfloat[\centering After pushing]{{\includegraphics[width=0.45\linewidth]{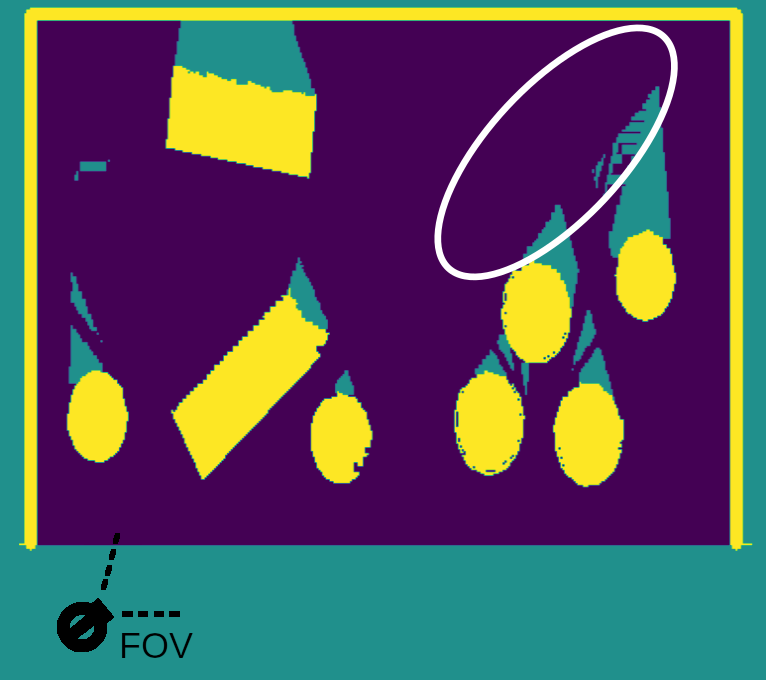} }}%
 \caption{Qualitative example of an occupancy map update in simulation after a
   minimally invasive push. (a) shows the map before the push~(green
   indicates unknown space), with the arrow indicating the planned
   push action. (b) shows the map after the push, where the cells in
   the marked area became visible due to the push.}
  \label{fig:push_advance}
\end{figure}


\begin{table}
\centering
{

\resizebox{\columnwidth}{!}{%
    \begin{tabular}{|c|c|c|}
\hline
\textbf{Metrics} &\textbf{Learned Push} & \textbf{Random Push}\\ \hhline{|=|=|=|}
entropy reduction & $27.62\% \pm  2.2$ & $\mathbf{37.05\%} \pm  2.6$\\ \hhline{|-|-|-|}
displacement & $^*\mathbf{0.24\,m} \pm  0.03$ & $0.34\, \textit{m} \pm  0.03$\\ \hhline{|-|-|-|}
\#iterations & $3.5 \pm  0.24$ &  $3.7  \pm  0.21$\\ \hhline{|-|-|-|}
object drop rate & $\mathbf{0.04}$& 0.21\\ \hhline{|-|-|-|}
\end{tabular}
}
}
\caption{Evaluation of the complete viewpoint push planner pipeline: Reduced entropy
  compared to the map before executing push actions, object
  displacement, and number of pipeline iterations as well as object drops. 
  Each metric is shown with its standard error.
  The experiments were carried out in 100 simulated scenarios.
  As can be seen, our complete pipeline was able to reduce the entropy of the map generated by our learned VPP even further.
  In comparison to a random push selection method, our trained push proposal network lead to reduced object displacements and a reduced object drop rate.
  The object displacements and entropy reduction are significant according to the paired t-test with a p-value of 0.05.
  }

\label{tab:Push_result}
\end{table}
To evaluate our complete pipeline and to show the influence of the push actions to reduce the entropy after VPP, we calculated the entropy before and after the push actions.
Furthermore, to show the benefits of our trained push prediction network, we applied a method that randomly selects a push candidate generated by our push sampling method~(see \secref{sec:data}).
Each combination of VPP followed by a push action is called a pipeline iteration and is terminated when the criteria defined in \secref{selection} is met.
We evaluated the pipeline on the same 100 simulated environments as used in \ref{sec:eval_vpp}.
As \tabref{tab:Push_result} shows, both pushing methods were clearly 
able to reduce the map entropy resulting from VPP alone by 27.62\% and 37.05\%,
with an average of 3.5 and 3.7 pipeline iterations.
While the random push selection method outperformed our trained method in terms of entropy reduction, our method shows the improvement wrt. the learned minimally invasive actions by reducing the likelihood of object drops and too large object displacements.
An object is considered dropped if it is tilted by more than $45^{\circ}$ or falls off the shelf.
For evaluating the displacement, we compared the object configuration before and after the full pipeline execution, using the Euclidean distance of the center point of each object as the evaluation metrics.
Out of the 100 simulated trials, our learned pipeline reduced the number of object drops after the push by 80\% and was able to reduce the displacement by 29.4\% compared to the random method.
To test the influence of the push length, we performed each trial with a push length of 10\,cm, 7\,cm, 5\,cm, and 2\,cm.
The longer the push length, the greater is the reduction in entropy.
However, also the displacements increase significantly according to the trade off between entropy reduction and configuration disruption.
Empirically, we evaluated 5\,cm to be the best in terms of this trade off and used it for the results in \tabref{tab:Push_result}.

Fig.
\ref{fig:push_advance} shows a qualitative comparison of a map before and after a push action.
As can be seen, the VPP is able to detect more, previously hidden space, while the push itself changed the object configuration only slightly.
On average, our method takes 0.31\,s to sample the push
candidates, while the network takes 0.1\,s for inference.

To summarize, we demonstrated that using a predicted non-prehensile push aids the VPP while maintaining the overall structure of the scene.
The complete pipeline is easily transferable to the real-robot system without loss of performance as can be seen in the accompanying video\footnote{https://www.youtube.com/watch?v=6C3Q2UFKq2Q}.


\subsection{Application of the Map Representation to Grasping}
\label{sec:exp_manipulation}
\begin{figure}[t]
  \centering
  \subfloat[\centering Object detection]{{\includegraphics[width=0.45\linewidth]{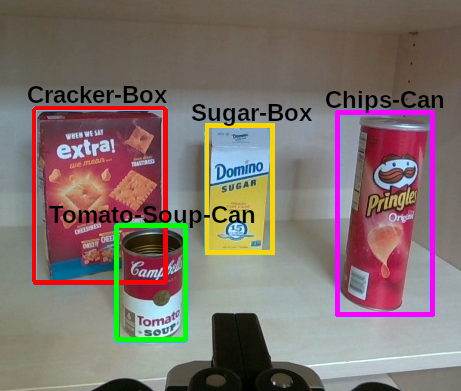} }}%
    \subfloat[\centering Labeld heightmap with grasp candidate]{{\includegraphics[width=0.45\linewidth]{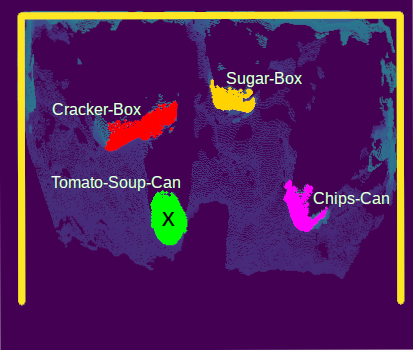} }}%
 \caption{Visualization of the capabilities of our map
   representation. The left image shows the current scene with the
   corresponding bounding boxes resulting from the object
   detection. On the right, our map representation is shown with the
   labels of each object added as a new layer and the center point of
   the tomato soup can for object retrieval.}
  \label{fig:detection}
\end{figure}

To demonstrate the benefit of our map representation, we performed two qualitative experiments on the real robot.
First, we added an additional object label layer to our pipeline.
To obtain the labels, we used an object detection network trained on the YCB dataset \cite{calli2015ycb} and transformed the object labels onto our map representation.
As shown in Fig.~\ref{fig:detection}, the object label layer can be used to reason about the individual objects in the scene.
We then used the map to grasp objects from the shelf, where we first applied a 2D~clustering on the labeled objects to obtain the center of each object cluster.
By deprojecting the corresponding pixel to its 3D Cartesian world coordinate with a predefined height value 10\,cm above the shelf board height, our system was able to grasp the soup can as well as the chips can in nine out of ten cases.
The single missed grasping trial was due to slippage of the object.

\section{Conclusion}
\label{sec:conclusion}
In this paper, we presented a novel pipeline for interactive viewpoint planning in confined spaces.
We demonstrated the efficacy of our approach in multiple simulated and real-world scenarios with a robotic arm equipped with an \mbox{RGB-D} camera on the end effector for mapping of the objects on a shelf.
The results show the improved performance of our system in comparison to several baselines and demonstrate that our agent is able to increase the mapped space.
We showed that the executed push actions highly benefit the viewpoint planner while being minimally invasive so that the overall object configuration can be maintained.
The runtime evaluation highlights the real-time capability of our system.
The code of our viewpoint push planner can be found on Github\footref{github_code}.

%


\bibliographystyle{IEEEtran}

\bibliography{bibliography}

\end{document}